\documentclass[sigconf]{acmart}
\usepackage{listings}
\usepackage{xcolor}
\usepackage{cleveref}

\renewcommand\footnotetextcopyrightpermission[1]{} % removes footnote with conference information in first column
\begin{document}
\newcommand{\mohamed}[1]{\textcolor{red}{#1}}
\newcommand{\anupam}[1]{\textcolor{blue}{#1}}
\newcommand{\anj}[1]{\textcolor{green}{#1}}

%\title{Quantifying Manifolds: Generative Models}
\title{Quantifying Manifolds: Do the manifolds learned by Generative Adversarial Networks converge to the real data manifold?}

\author{Anupam Chaudhuri, Anj Simmons, Mohamed Abdelrazek}
\email{ anupam.chaudhuri@deakin.edu.au, a.simmons@deakin.edu.au, mohamed.abdelrazek@deakin.edu.au}
\affiliation{\institution{Applied Artificial Intelligence Institute} \state{VIC} \country{Australia}}

\begin{abstract}
This paper presents our experiments to quantify the manifolds learned by ML models (in our experiment, we use a GAN model) as they train.
%We compare these epoch manifolds to the real manifolds representing the real data.
We compare the manifolds learned at each epoch to the real manifolds representing the real data.

%To quantify a manifold, studied what metrics we extract from the manifold learned by the ML model, how these metrics change as we continue to train the model, and how we can measure the convergence of the model using manifold metrics.

To quantify a manifold, we study the intrinsic dimensions and topological features of the manifold learned by the ML model, how these metrics change as we continue to train the model, and whether these metrics convergence over the course of training to the metrics of the real data manifold.

%We believe the same set of metrics can be used to compare two models - i.e. are two ML models equivalent? overlap? etc. Furthermore, these metrics can inform the training process as a new cost function.
\end{abstract}
\maketitle
% https://tex.stackexchange.com/questions/522971/how-can-i-remove-the-running-header-from-acm-sigconf-template
\pagestyle{plain}

\section{Introduction}
\label{sec:introduction}
Robustness is one of the top challenges facing the ML field - i.e. how could we make sure that the output of the ML model is correct, and how to quantify the uncertainties in the model output. Existing techniques address this problem from different perspectives including in and out of distribution, guards, interpretability, and chain of thoughts. Yet, there is no profound solution. These techniques are usually model-specific.

This paper introduces another perspective to the robustness problem focused on the training data. Our hypothesis is that by understanding and quantifying the geometry of data manifolds and the ML learned manifolds, we can gain deeper insights into the robustness of ML models.

A manifold is a topological space that locally resembles Euclidean space, but globally may have a more complex structure. % Manifold theory aims to understand how complex, high-dimensional structures can be represented in a lower-dimensional space. This motivated multiple manifold learning methods - e.g. t-SNE, UMAP, Isomap - as a dimensionality reduction techniques.
% During the training process, for example training a GAN model, the generator learns to map points from a low-dimensional latent space to the high-dimensional manifold of the real-data. 
As the model learns, the learned manifold is expect to approximate the manifold of the real data distribution. The key question is how do we quantify a manifold (what matrics should we consider), and how these metrics evolve during the training process of the ML model. We have identified a suite of metrics that we use to quantify the ML learned manifold at each epoch including the number of intrinsic dimensions, and topological features $H_0$, $H_1$, and $H_2$.

\section{Topological Data analysis}
\label{sec:tda}
Topology deals with the properties of geometric objects that are invariant under continuous deformation. For example, a donut and a coffee cup are considered topologically the same object, and similarly, a point and a ball are topologically similar. The branch of algebraic topology attaches algebraic objects such as groups to these topological spaces, as mentioned in Hatcher's work~\cite{hatcher2001alge}. This forms an identification or features of the topological space.
%Merkulov's work~\cite{merkulov2003hatcher}

Now using the idea of persistence; we can extend the topology to the finite data sets. The current from of persistence tools can be traced back to~\cite{frosini1992measuring, robins1999towards}. Homology groups form the essential invariants for a Topological object. Hence persistence homology groups form the invariant for finite discrete objects. 

\subsubsection{Simplicial complex and Simplicial Chain}
In mathematics, it is a common practice to study simple objects that can be visualized or computed and then approximate complex objects in relation to the known or studied objects. Therefore, we introduce the concept of a simplicial complex, which facilitates the computational aspects of algebraic topology.

Given a set $X = \{x_0, \ldots, x_k\} \subset \mathbb{R}^d$ of $k + 1$ affinely independent points, the $k$-dimensional simplex $\sigma = [x_0, \ldots, x_k]$ spanned by $X$ is the convex hull of $X$. The points of $X$ are called the vertices of $\sigma$, and the simplices spanned by the subsets of $X$ are called the faces of $\sigma$. A geometric simplicial complex $K$ in $\mathbb{R}^d$ is a collection of simplices such that:
\begin{enumerate}
\item 
Any face of a simplex of $K$ is a simplex of $K$,
\item 
The intersection of any two simplices of $K$ is either empty or a common face of both.
\end{enumerate}

Let's begin with an example of a geometrical simplex 

\begin{align*}
\text{Let } K &= \{\left[a\right], \left[b\right], \left[c\right], \\
&\quad  \left[a, b\right], \left[b, c\right], \left[a, c\right], \left[a, b, c\right]\}.
\end{align*}

The above is a simplicial complex of dimension two as the highest simplex $\left[a, b, c\right]$ is of dimension two. 

For each dimension \(k\), there is a chain group \(C_k\) associated with the \(k\)-dimensional simplices. The chain group is the free abelian group generated by the \(k\)-dimensional simplices of the simplicial complex.

Valid 1-simplex for $K$ which is an element of the chain group $C_1$ is given by
\begin{itemize}
\item $\left[a, b\right]$
\item $\left[a, b\right] + \left[a, c\right]$
\item \textit{and any other similar combination of 1-dimensional simplex.}
\end{itemize}

%\subsubsection{Boundary homomorphism} ~\\
%
\textbf{Boundary homomorphism}
Let $\sigma = \left\{v_0, \dots, v_p \right\}$ be a simplicial complex of dimension $p$. We define a boundary homomorphism $\partial \sigma = \sum_i v_0, \dots, \hat{v_i}, \dots, v_p$ where $\hat{v_i}$ denote it has been eliminated from the set. Hence we get a map from 
$\partial_p : C_p \to C_{p-1}$. 
Let's take a particular example 
\begin{align}
\partial_2 \{a,b,c\} &= \{b,c\} + \{a,c\} + \{a,b\} \\
\partial_1(\{b,c\} + \{a,c\} + \{a,b\}) &= \{c\} + \{b\} + \{c\} + \{a\} + \{b\} + \{a\} \\
        & = 0\quad \text{if we calculate in $\mathbb{Z}_2$}
\end{align}

\textbf{Chain Complex}

A $k$ dimensional simplex forms a vector space, which we denote as $C_k$

\begin{align*}
0 \rightarrow C_n \xrightarrow{\partial_{n+1}} C_{n-1} \xrightarrow{\partial_{n}} \cdots \rightarrow C_1 \xrightarrow{\partial_2} C_0 \xrightarrow{\partial_1} 0
\end{align*}

\textbf{Diagram}

\textit{Kernel vs image}

\begin{align*}
f: & A \rightarrow B \\
\text{ker } f &= \{ a \in A \mid f(a) = 0\} \\
\text{im } f &= \{ f(a) \mid a \in A\} \subseteq B
\end{align*}

\textbf{Cycle and boundary groups}

\begin{align*}
\text{Cycle group } Z_p &= \text{ker } \partial_p \\
\text{Boundary group } B_p &= \text{im } \partial_{p+1}
\end{align*}

% \textbf{Chain Complex,}

% \begin{align*}
% 0 \rightarrow C_n \xrightarrow{\partial_{n+1}} C_{n-1} \xrightarrow{\partial_{n}} \cdots \rightarrow C_1 \xrightarrow{\partial_2} C_0 \xrightarrow{\partial_1} 0
% \end{align*}

\[
B_p \subset Z_p
\]

\begin{definition}
 Simplicial homology group and Betti numbers the $k^{\text{th}}$ (simplicial) homology group of $K$ is the quotient vector space
\[
H_k(K) = \frac{Z_k(K)}{B_k(K)}.
\]
The $k^{\text{th}}$ Betti number of $K$ is the dimension $\beta_k(K) = \dim H_k(K)$ of the vector space $H_k(K)$.
\end{definition}

Hence given a simplicial complex we can compute the Homology group. Now given a finite set of data points $X$:
\[ X = \{ p_1, p_2, \ldots, p_m \} \]
and a matrix that gives the distance of the points (or equivalently if the points are embedded in a metric space and a metric dist such as the Euclidean distance), pick a threshold \(\varepsilon\):
\[ \mathcal{V}_\varepsilon(X) := \{ \sigma \subset X | \forall u,v \in \sigma, \text{dist}(u,v) \leq \varepsilon \} \]

$\mathcal{V}_\varepsilon(X)$ is known as Vietoris Rips complex. Given $\varepsilon_1$ and $\varepsilon_2$ such that $\varepsilon_1\leq \varepsilon_2$ then 

\[
\mathcal{V}_{\varepsilon_1 (X)} \subseteq \mathcal{V}_{\varepsilon_2}(X)
\]

\begin{definition}{Filtration}\\
    Given a set or space \(X\), a \emph{filtration} \(\mathcal{F}\) of \(X\) is a sequence of subspaces or subsets:
\[ X_0 \subseteq X_1 \subseteq X_2 \subseteq \dots \subseteq X_n = X \]
Each \(X_i\) is referred to as the \(i\)-th filtration level or step of the filtration.

\end{definition}

So as we vary $\varepsilon$ we get a filtration of Vietoris Rips complex.  
So in general for a topological space $X$ given a filtration of $X$
\[ 
 K_i \subseteq K_j \subseteq X
\]
 
\[
f^{i,j}_p:H_p(K_i) \xrightarrow H_p(K_j)
\]

\begin{align*}
  &  0 \to H_p(K_0) \xrightarrow{f^{0,1}_p} H_p(K_1) \xrightarrow{f^{1,2}_p} \dots \xrightarrow{f^{n-2,n-1}_p} H_p(K_{n-1})\\ 
  & \xrightarrow{f^{n-1,n}_p} H_p(K_n) \to H_p(X)
\end{align*}

The Persistence homology group is then computed by 
\[
H_p^{i,j} = \frac{Z_p (K_i)}{B_p (K_j)\cap Z_p (K_i)}
\]

So particular to our context for a monotonic sequence of $\varepsilon_i$. 
\[
\varepsilon_1 \leq \varepsilon_2, \ldots, \leq \varepsilon_i \ldots \leq \varepsilon_j \ldots
\]
we get an filtration 
\begin{equation} 
\label{filtration}
\mathcal{VR}:\mathcal{V}_{\varepsilon_1 (X)} \subseteq \mathcal{V}_{\varepsilon_2}(X)\ldots \mathcal{V}_{\varepsilon_i (X)}\ldots \subseteq \mathcal{V}_{\varepsilon_j}(X)
\end{equation}
As it's clear we are working in topological space $X$ we will denote $\varepsilon_{i}(X)$, by $\varepsilon_i$.

\[
H_k(\iota_{i,j}):H_k(V_{\varepsilon_i}) \to H_k(V_{\varepsilon_j})
\]

A homology class $\alpha \in H_k(V_{\varepsilon_i})$ is said to be born at $V_{\varepsilon_i}$ if $\alpha \notin H^{i-1, i}_k$, i.e., if it is not in the image of $H_k(\iota_{i-1, i})$. 

If $\alpha$ is born at $V_{\varepsilon_i}$, it is said to die at $V_{\varepsilon_j}$ if $H_k(\iota_{i,j-1})(\alpha) \notin H^{i-1,j-1}_k$ and $H^k(\iota_{i,j})(\alpha) \in H^{i-1,j}_k$. 

The persistence of $\alpha$ is given by $\varepsilon_j - \varepsilon_i$ and set to infinity if it never dies. 

The persistent Betti-numbers, defined by $\beta_{i,j}^k := \text{dim} H^{i,j}_k$, carry information on how the homology (and thus the topology) changes across the filtration.

\subsection{Persistence Diagram}
Given a filtration~\cref{filtration} by varying $\epsilon_{i} $ we define the persistence diagram for the 
$k$-th homology group as the set $PD_{k}(\mathcal{VR})$ to be 

\begin{equation}
    PD_{k}(\mathcal{VR}): = \left\{\left( b_i , d_i\right)\mid i\in \mathbb{N}, b_i , d_i \in \{ \epsilon_0, \ldots, \epsilon_n,\ldots\} \right\}
\end{equation}
where $b_i$ in the tuple $\left( b_i , d_i\right)$ records the appearance of a $k$ dimensional homology group and $d_i$ records it's disapperance. In the event if the homology group persists, i.e. it does not disappear during the end of filtration, we set $d_i = \infty$.

\section{Categorial framework to understand the quantification}
Given a point cloud we want to associate an set of object  in the category of filtered topological spaces. 

\textbf{Category of Filtrated Topological Spaces:}

Let $\mathcal{F}$ be the category of filtrated topological spaces. An object in $\mathcal{F}$ is a pair $(X, \mathcal{F})$, where $X$ is a topological space and $\mathcal{F}$ is a filtration of $X$.

A morphism from $(X, \mathcal{F})$ to $(Y, \mathcal{G})$ is a continuous map $f: X \rightarrow Y$ such that $f(X_t) \subseteq Y_t$ for all $t$.

\textbf{Persistence Diagram Functor:}

The persistence diagram is a functor $\text{PD}: \mathcal{F} \rightarrow \mathcal{D}$, where $\mathcal{D}$ is the category of persistence diagrams and their morphisms.

\textbf{On Objects:} For each object $(X, \mathcal{F})$ in $\mathcal{F}$, $\text{PD}(X, \mathcal{F})$ is the associated persistence diagram.

\textbf{On Morphisms:} For a morphism $f: (X, \mathcal{F}) \rightarrow (Y, \mathcal{G})$ in $\mathcal{F}$, $\text{PD}(f): \text{PD}(X, \mathcal{F}) \rightarrow \text{PD}(Y, \mathcal{G})$ is a map respecting the birth and death of persistent features.

\textbf{Functorial Properties:}

1. \textbf{Identity:} For any object $(X, \mathcal{F})$, $\text{PD}(\text{id}): \text{PD}(X, \mathcal{F}) \rightarrow \text{PD}(X, \mathcal{F})$ is the identity map.

2. \textbf{Composition:} If $f: (X, \mathcal{F}) \rightarrow (Y, \mathcal{G})$ and $g: (Y, \mathcal{G}) \rightarrow (Z, \mathcal{H})$ are morphisms, then $\text{PD}(g \circ f) = \text{PD}(g) \circ \text{PD}(f)$.

For the sake of comparison we can attach number to the object in the category of persistence diagram. This can be done by using entropy, Wasserstein distance, entropy, etc. %amplitude etc.

\textbf{Metrics:}

1. \textbf{Wasserstein Distance for Persistence Diagrams:}
The $p$-Wasserstein distance between two persistence diagrams $D_1$ and $D_2$ is the infimum over all bijections $\gamma$ of

\[
W_p(D_1, D_2) = \left( \inf_{\gamma} \sum_{(x, y) \in D_1 \times D_2} \lVert x - \gamma(y) \rVert^p \right)^{1/p}
\]

where $\lVert x - \gamma(y) \rVert$ is the distance between the points $x$ and $\gamma(y)$, and $p \geq 1$ is a parameter that determines the order of the Wasserstein distance.

The limit as $p$ approaches infinity defines the bottleneck distance. More explicitly, it is the infimum over the same set of bijections of the value

\[
W_{\infty}(D_1, D_2) = \inf_{\gamma} \sup_{(x, y) \in D_1 \times D_2} \min\{\lVert x - \gamma(y) \rVert, \lVert x - \gamma(y) \rVert\}
\]

In this paper, we study the Wasserstein distance from the trivial diagram.

2. \textbf{Entropy for Persistence Diagrams:}
This feature quantifies the complexity of the persistence diagram as calculated by the Shannon entropy of the persistence values (birth and death), with higher entropy indicating a more complex topology.

% \textbf{Metric space:}

% The set of persistence diagrams together with any of the distance metrics defined above forms a metric space.

\section{Manifold Learning}
\label{sec:manifolds}
The concept of a manifold traces back to the 19th century, with contributions from mathematicians like Bernhard Riemann, who introduced Riemannian manifolds in his 1854 habilitation lecture. This period marked the beginning of manifold theory, initially motivated by problems in mathematical analysis and the need to generalize concepts of curves can be consider as one dimensioanl manifold and surfaces which can be considered as two dimensional manifold to higher dimensions. Over time, the manifold concept evolved to become a foundational element in various branches of mathematics and physics, influencing the development of topology, differential geometry, and the theory of relativity.

\begin{definition}[Manifold]
A manifold \(M\) of dimension \(n\) is a topological space that satisfies the following properties:
\begin{enumerate}
    \item \textbf{Locally Euclidean (of dimension \(n\))}: For every point \(x \in M\), there exists an open neighborhood \(U\) of \(x\) that is homeomorphic to an open subset of \(\mathbb{R}^n\). This means there is a continuous bijection with a continuous inverse between \(U\) and an open subset of \(\mathbb{R}^n\).
    \item \textbf{Hausdorff and Second Countable}:
    \begin{itemize}
        \item \textbf{Hausdorff}: For any two distinct points \(x, y \in M\), there exist disjoint open neighborhoods \(U\) of \(x\) and \(V\) of \(y\).
        \item \textbf{Second Countable}: There exists a countable basis for the topology of \(M\).
    \end{itemize}
\end{enumerate}
These conditions ensure that manifolds are spaces that are locally like Euclidean space while potentially having complex global structures.
\end{definition}

%These things are mathematically well defined for a continous object, but when we are dealing with discrete objects we need to estimate them using methods from Topological Data Analysis.

\section{Manifold Quantification Metrics}
\label{metrics}
With the introduction of manifolds now we can do calculus over compact object like Sphere, torus or smooth objects. Hence we can quantify manifold by measuring it's dimension, topological invariants, like homology, etc.

\subsection{Intrinsic dimensions}
Estimating the intrinsic dimension (ID) holds significant importance in the field of machine learning, extending beyond the scope of dimensionality reduction techniques to address a variety of critical issues. Primarily, employing dimensions beyond what is essential can introduce several challenges, including increased storage requirements for data and a reduction in algorithmic performance, which typically scales with the dimensionality of the data. Furthermore, as the number of dimensions escalates, the difficulty of constructing dependable classifiers intensifies, a phenomenon known as the curse of dimensionality~\cite{carleo2019machine}.

One of the most intuitive definition of intrinsic dimension is given by Minkowski–Bouligand dimension

\begin{definition}
    The Minkowski–Bouligand dimension, also referred to as the box-counting dimension, is a measure used to quantify the fractal dimension of a set $S$ within a Euclidean space. It provides an assessment of the complexity or "roughness" of $S$ by examining how the number of small boxes, required to cover the set, scales with the size of these boxes.

Formally, given a set $S$, the Minkowski–Bouligand dimension $D$ is defined as:

\begin{equation}
D = \lim_{\epsilon \to 0} \frac{\log N(\epsilon)}{\log(1/\epsilon)}
\end{equation}

where:
\begin{itemize}
    \item $N(\epsilon)$ denotes the minimum number of boxes of side length $\epsilon$ needed to completely cover the set $S$.
    \item $\epsilon$ is the side length of the boxes, approaching zero.
\end{itemize}
\end{definition}
This dimension reflects the rate at which the detail or complexity of the set $S$ increases as the scale of observation becomes finer, providing insight into the fractal nature of the set.
Intrinsic dimension as computed in the paper~\cite{glielmo2022dadapy}

In the case of an integer number of intrinsic dimensions, they can be thought of as the minimum number of parameters needed to describe a point on a manifold. For example, even if a 2D surface such as a sheet of paper is rolled up like a swiss roll, only two intrinsic dimensions are required to describe a point on the surface despite the surface being embedded in 3D space.

Images generated by a GAN can be thought of as sampling datpoints that lie on a manifold. Our goal is to estimate the intrinsic dimensions of this manifold, given only a set of datapoints (a pointcloud) sampled from the manifold. In this paper, we use the two nearest neigbour (2NN) method \cite{facco2017estimating} implemented in dadapy\footnote{\url{https://dadapy.readthedocs.io/en/latest/id_estimation.html}}~\cite{dadapy} to estimate the instrinsic dimension of the dataset.

Previous work has studied the intrinsic dimensions of data representations at each layer of convolutional neural networks, finding that the estimated intrinsic dimensions increase during the initial layers and decrease during the final layers \cite{ansuini2019intrinsic}. Furthermore, the intrinsic dimensions of the last hidden layer was found to be predictive of generalisation performance \cite{ansuini2019intrinsic}. Recent work~\cite{magai2022topology,magai2023deep} has also used the persistent homological fractal dimension~\cite{schweinhart2020fractal} as a more accurate estimate of intrinsic dimensions and studied this in the context of convolutional neural networks.

\subsection{Topological Features}

Topological features characterise a pointcloud based on how the number of connected components ($H_0$), holes ($H_1$) and voids ($H_2$) vary as the function of a threshold $\epsilon_i$ controlling the distance between which points are considered connected. A formal definiton is given in \autoref{sec:tda}. In this work, we study $H_0$, $H_1$ and $H_2$ using the metrics of entropy and Wasserstein distance from the trivial diagram. We extract these topological features using the Topological Data Analysis (TDA) Python library giotto-tda\footnote{\url{https://giotto-ai.github.io/gtda-docs/}}~\cite{giotto-tda}.

%  (which we prove is proportional to the sum of the lifespans in \autoref{appendix:wasserstein-proportional-lifespan-sum})

Recent work~\cite{magai2022topology,magai2023deep} has studied the sum of the lifespans in $H_0$ and $H_1$ for convolutional neural networks, specifically how this varies in each layer of the network and also how it varies with the training epoch. Recent work has also proposed using a topological feature vector (specifically lifespans in $H_0$ of a pointcloud of image embeddings) to evaluate the quality of a GAN \cite{horak2021topology} by measuring the distance between the topological feature vector of the generated distribution of images and the topological feature vector of the real distribution of images.

\subsection{Gaps}

While previous work has examined intrinsic dimensions and topological features of neural networks, these have either focussed on convolutional neural networks \cite{ansuini2019intrinsic,magai2022topology,magai2023deep} or topological features of GANs in $H_0$ only \cite{horak2021topology}. To the best of our knowledge, we are the first to propose a comprehensive evaluation of multiple manifold quanficiation metrics to gain insights into the GAN learning process.

\section{Experiment}
To train a Generative Adversarial Network (GAN), we begin with a set of images, which we will refer to as the `real image set', denoted by $\mathcal{I}{r}$. For the sake of simplicity, let us assume that each image in $\mathcal{I}{r}$ has dimensions of $32 \times 32 \times 3$. Consequently, we can embed each image into a vector space, $\mathbb{R}^{32 \times 32 \times 3}$. This embedding is achieved by considering each pixel as a dimension or feature. It should be noted that there are numerous other, more efficient methods for embedding images. Given that $32 \times 32 \times 3 = 3072$, each image is thus represented as a point in a 3072-dimensional space. 

For the real image set, we obtain a point cloud representation. Given that this representation forms a point cloud, we can apply Topological Data Analysis (TDA) to characterize it with features such as persistence homology diagrams, denoted as $PD_{k}^{r}$. Since the persistence homology dimension constitutes a set, we can quantify this set by employing measures such as entropy or the Wasserstein distance. For instance, the entropy associated with $PD_{k}^{r}$ can be represented as $E_{k}^{r}$.

We hypothesize that the performance of a Generative Adversarial Network (GAN) can be evaluated by examining the point cloud for the generated images at the $i$th epoch, which we denote as $\mathcal{I}{g_i}$. The corresponding persistence diagram is denoted as $PD{k}^{g_i}$. Similarly, we can associate an entropy value with this, which we denote as $E_{k}^{g_i}$. This approach allows us to quantitatively assess the GAN's performance in generating images that are topologically consistent with the real image set.

We hypothesise that this convergence should also hold for other metrics that quantify the manifold such as the number of intrinsic dimensions.

\subsection{Dataset and Model}

We perform a preliminary evaluation of our approach applied to a GAN trained on cat images from the CIFAR-10 dataset. The architcture of the GAN follows a popular online tutorial\footnote{\url{https://machinelearningmastery.com/how-to-develop-a-generative-adversarial-network-for-a-cifar-10-small-object-photographs-from-scratch/}}. The GAN implementation (adapted to generating cat images) and our code for extracting metrics is publically available online\footnote{\url{https://github.com/Anugithub2019/GAN}}.

We provide a summary of the architecure 

\begin{itemize}
    \item[1.] \textbf{Data Loading and Visualization}

    CIFAR-10 dataset is loaded, and cat images are extracted. The first 49 images from the training set are visualized.

    \item[2.] \textbf{Discriminator Model}

    The discriminator is a convolutional neural network (CNN) that takes (32, 32, 3) images as input. It uses LeakyReLU activations, down-sampling, dropout for regularization, and outputs binary classification results. Binary cross-entropy loss and the Adam optimizer are used for training.

    \item[3.] \textbf{Generator Model}

    The generator is a neural network that takes a noise vector of size 100 as input. Dense layers upscale the noise into a 3D tensor, and transpose convolutional layers (Conv2DTranspose) up-sample the data. The output layer uses a tanh activation function to generate pixel values in the range [-1, 1].

    \item[4.] \textbf{GAN Model}

    The GAN model combines the generator and discriminator. The discriminator's weights are set as non-trainable. The model is compiled with binary cross-entropy loss and the Adam optimizer.

    \item[5.] \textbf{Training Process}

    Real and fake samples are generated for training the discriminator. The discriminator and generator models are trained iteratively. The discriminator is trained on both real and fake samples, and the generator is trained using the GAN model with inverted labels. Model performance and generated images are periodically summarized during training.

    \item[6.] \textbf{Training Loop}

    The training loop runs for 500 epochs with a batch size of 128. Performance and results are periodically summarized.

\end{itemize}

\section{Results}
\label{evaluation}

\begin{figure*}
    \centering
    \includegraphics[width=0.5\linewidth]{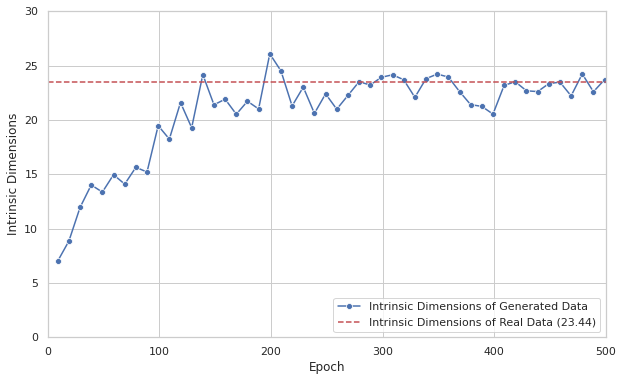}
    \caption{Intrinsic Dimension vs Epoch}
    \label{fig:id-vs-epoch}
\end{figure*}

We observed in \autoref{fig:id-vs-epoch} that the intrinsic dimensions of the manifold over which images are generated starts with a lower number of intrinsic dimensions than the real image distribution, and then over the course of training appears to converge towards the intrinsic dimensions of the real image distribution. Even in the case of the real image distribution, the estimated intrinsic dimensions (approx. 23 dimensions) is far lower than the total number of dimensions (3072 dimensions), confirming the mainfold hypothesis that the real world data lies on a low dimensional manifold embedded in a high dimensional space.

\begin{figure*}
    \centering
    \includegraphics[width=0.45\linewidth]{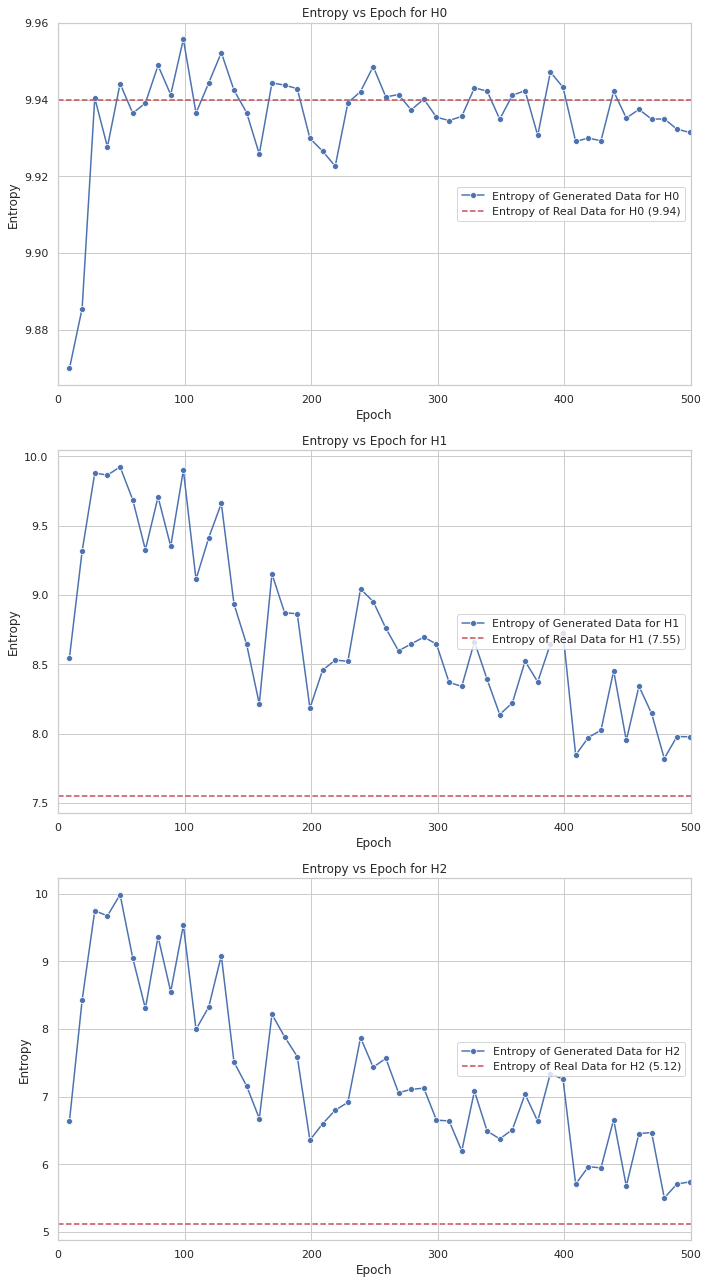}
    \includegraphics[width=0.45\linewidth]{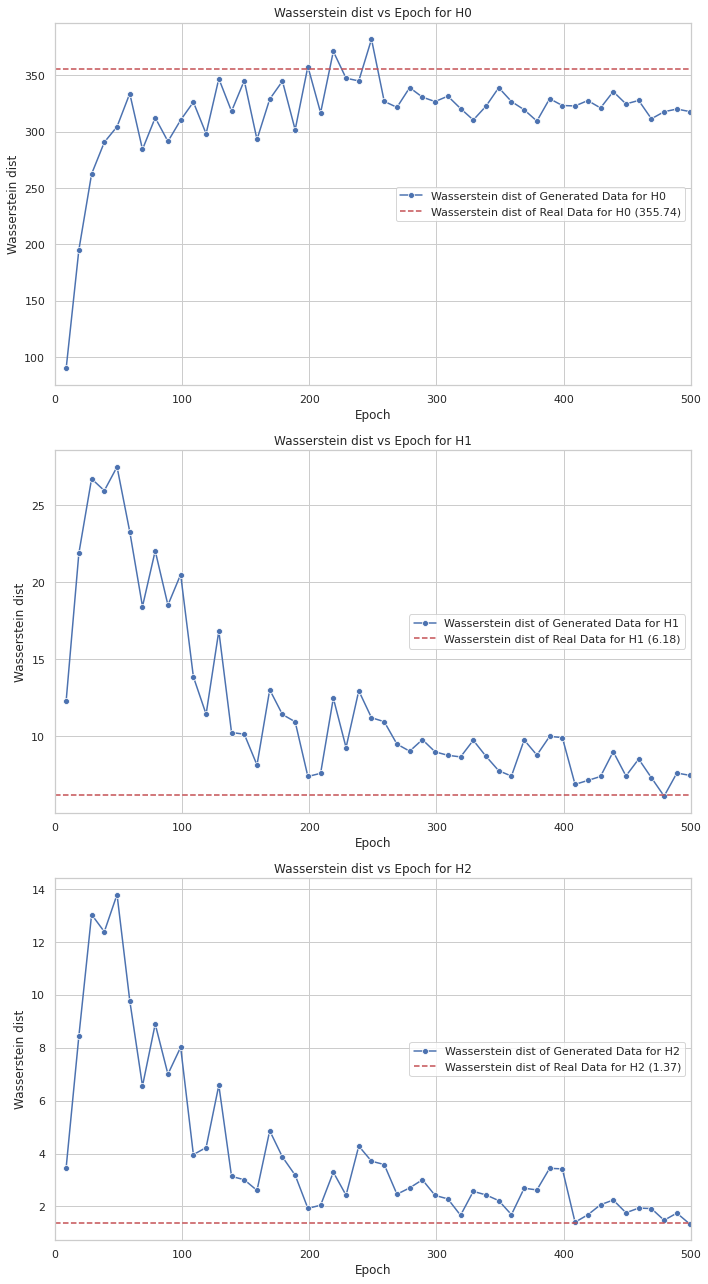}
    \caption{Topological Features vs Epoch}
    \label{fig:topology}
\end{figure*}

For the topological features in \autoref{fig:topology}, we observe that both the Entropy and Wasserstein distance metrics for the generated data converge towards that of the real data. However, $H_0$ appears to converge faster than $H_1$ and $H_2$.

%\section{Discussion}
%\label{discussion}

\section{Conclusion}
\label{conclusion}

Our quantification of the manifolds from which GANs generate data provides insights into how the manifolds evolve over the course of training. Specifically, the dimensions and topology of the manifolds does not match the real data distribution at first, but converges towards that of the real data training process. We also found that intrinsic dimensions and $H_0$ appears to converge towards that of the real data faster than $H_1$ and $H_2$. Further research is needed to see if these patterns generalise to other GANs and generative models and qualitatively assess how this impacts the percieved quality of the generated images. 

\bibliography{references.bib}

\begin{thebibliography}{13}
\providecommand{\natexlab}[1]{#1}
\providecommand{\url}[1]{\texttt{#1}}
\expandafter\ifx\csname urlstyle\endcsname\relax
  \providecommand{\doi}[1]{doi: #1}\else
  \providecommand{\doi}{doi: \begingroup \urlstyle{rm}\Url}\fi

\bibitem[Ansuini et~al.(2019)Ansuini, Laio, Macke, and Zoccolan]{ansuini2019intrinsic}
Alessio Ansuini, Alessandro Laio, Jakob~H Macke, and Davide Zoccolan.
\newblock Intrinsic dimension of data representations in deep neural networks.
\newblock \emph{Advances in Neural Information Processing Systems}, 32, 2019.

\bibitem[Carleo et~al.(2019)Carleo, Cirac, Cranmer, Daudet, Schuld, Tishby, Vogt-Maranto, and Zdeborov{\'a}]{carleo2019machine}
Giuseppe Carleo, Ignacio Cirac, Kyle Cranmer, Laurent Daudet, Maria Schuld, Naftali Tishby, Leslie Vogt-Maranto, and Lenka Zdeborov{\'a}.
\newblock Machine learning and the physical sciences.
\newblock \emph{Reviews of Modern Physics}, 91\penalty0 (4):\penalty0 045002, 2019.

\bibitem[Facco et~al.(2017)Facco, d’Errico, Rodriguez, and Laio]{facco2017estimating}
Elena Facco, Maria d’Errico, Alex Rodriguez, and Alessandro Laio.
\newblock Estimating the intrinsic dimension of datasets by a minimal neighborhood information.
\newblock \emph{Scientific reports}, 7\penalty0 (1):\penalty0 12140, 2017.

\bibitem[Frosini(1992)]{frosini1992measuring}
Patrizio Frosini.
\newblock Measuring shapes by size functions.
\newblock In \emph{Intelligent Robots and Computer Vision X: Algorithms and Techniques}, volume 1607, pages 122--133. SPIE, 1992.

\bibitem[Glielmo et~al.(2022{\natexlab{a}})Glielmo, Macocco, Doimo, Carli, Zeni, Wild, d’Errico, Rodriguez, and Laio]{dadapy}
Aldo Glielmo, Iuri Macocco, Diego Doimo, Matteo Carli, Claudio Zeni, Romina Wild, Maria d’Errico, Alex Rodriguez, and Alessandro Laio.
\newblock Dadapy: Distance-based analysis of data-manifolds in python.
\newblock \emph{Patterns}, page 100589, 2022{\natexlab{a}}.
\newblock ISSN 2666-3899.
\newblock \doi{https://doi.org/10.1016/j.patter.2022.100589}.
\newblock URL \url{https://www.sciencedirect.com/science/article/pii/S2666389922002070}.

\bibitem[Glielmo et~al.(2022{\natexlab{b}})Glielmo, Macocco, Doimo, Carli, Zeni, Wild, d’Errico, Rodriguez, and Laio]{glielmo2022dadapy}
Aldo Glielmo, Iuri Macocco, Diego Doimo, Matteo Carli, Claudio Zeni, Romina Wild, Maria d’Errico, Alex Rodriguez, and Alessandro Laio.
\newblock Dadapy: Distance-based analysis of data-manifolds in python.
\newblock \emph{Patterns}, 3\penalty0 (10), 2022{\natexlab{b}}.

\bibitem[Hatcher(2001)]{hatcher2001alge}
Allen Hatcher.
\newblock Alge braic topology, 2001.

\bibitem[Horak et~al.(2021)Horak, Yu, and Salimi-Khorshidi]{horak2021topology}
Danijela Horak, Simiao Yu, and Gholamreza Salimi-Khorshidi.
\newblock Topology distance: A topology-based approach for evaluating generative adversarial networks.
\newblock In \emph{Proceedings of the AAAI Conference on Artificial Intelligence}, volume~35, pages 7721--7728, 2021.

\bibitem[Magai(2023)]{magai2023deep}
German Magai.
\newblock Deep neural networks architectures from the perspective of manifold learning.
\newblock \emph{arXiv preprint arXiv:2306.03406}, 2023.

\bibitem[Magai and Ayzenberg(2022)]{magai2022topology}
German Magai and Anton Ayzenberg.
\newblock Topology and geometry of data manifold in deep learning.
\newblock \emph{arXiv e-prints}, pages arXiv--2204, 2022.

\bibitem[Robins(1999)]{robins1999towards}
Vanessa Robins.
\newblock Towards computing homology from finite approximations.
\newblock In \emph{Topology proceedings}, volume~24, pages 503--532, 1999.

\bibitem[Schweinhart(2020)]{schweinhart2020fractal}
Benjamin Schweinhart.
\newblock Fractal dimension and the persistent homology of random geometric complexes.
\newblock \emph{Advances in Mathematics}, 372:\penalty0 107291, 2020.

\bibitem[Tauzin et~al.(2021)Tauzin, Lupo, Tunstall, P\'{e}rez, Caorsi, Medina-Mardones, Dassatti, and Hess]{giotto-tda}
Guillaume Tauzin, Umberto Lupo, Lewis Tunstall, Julian~Burella P\'{e}rez, Matteo Caorsi, Anibal~M. Medina-Mardones, Alberto Dassatti, and Kathryn Hess.
\newblock giotto-tda: A topological data analysis toolkit for machine learning and data exploration.
\newblock \emph{Journal of Machine Learning Research}, 22\penalty0 (39):\penalty0 1--6, 2021.
\newblock URL \url{http://jmlr.org/papers/v22/20-325.html}.

\end{thebibliography}
\bibliographystyle{plainnat}

\appendix
\begin{comment}
\section{Proofs}
\subsection{Proof that Wasserstein distance from the trivial diagram is proportional to the sum of the lifespans}
\label{appendix:wasserstein-proportional-lifespan-sum}
\end{comment}

\end{document}